%% file: main.tex
\documentclass[final]{cvpr}

\usepackage{times}
\usepackage{epsfig}
\usepackage{graphicx}
\usepackage{amsmath}
\usepackage{amssymb}
\usepackage{comment}
\usepackage{multirow}
\usepackage[pagebackref=true,breaklinks=true,colorlinks,bookmarks=false]{hyperref}

\def\cvprPaperID{****} % *** Enter the CVPR Paper ID here
\def\confYear{CVPR 2021}
\begin{document}

%%%%%%%%% TITLE
\title{Video Surveillance for Road Traffic Monitoring}

\author{Pol Albacar \quad \`Oscar Lorente \quad Eduard Mainou \quad Ian Riera \\
Universitat Polit\`ecnica de Catalunya\\
{\tt\small \{pol.albacar \quad oscar.lorente \quad eduard.mainou \quad ian.pau.riera\}@estudiantat.upc.edu}
% \and
}

\maketitle

%%%%%%%%% ABSTRACT
%The ABSTRACT is to be in fully-justified italicized text, at the top
%   of the left-hand column, below the author and affiliation
%   information. Use the word ``Abstract'' as the title, in 12-point
%   Times, boldface type, centered relative to the column, initially
%   capitalized. The abstract is to be in 10-point, single-spaced type.
%   Leave two blank lines after the Abstract, then begin the main text.
%   Look at previous CVPR abstracts to get a feel for style and length.
\begin{abstract}
  This paper presents the learned techniques during the Video Analysis Module of the Master in Computer Vision from the Universitat Autònoma de Barcelona, used to solve the third track of the AI-City Challenge~\cite{aic21}. This challenge aims to track vehicles across multiple cameras placed in multiple intersections spread out over a city. The methodology followed focuses first in solving multi-tracking in a single camera and then extending it to multiple cameras. The qualitative results of the implemented techniques are presented using standard metrics for video analysis such as mAP for object detection and IDF1 for tracking.
The source code is publicly available at:
\url{https://github.com/mcv-m6-video/mcv-m6-2021-team4}.

\end{abstract}

%%%%%%%%% BODY TEXT

% Motivation
\input{src/1_motivation}

% Related Work
\input{src/2_related_work}

% Methodology
\input{src/3_methodology}

% Evaluation / Results
\input{src/4_evaluation}

% Conclusions
\input{src/5_conclusions}

{\small
\bibliographystyle{ieeetr}
\bibliography{egbib}
}

\end{document}

%% file: src/1_motivation.tex
\section{Motivation}
The aim of this project is to solve the CVPR 2021 AI City Challenge~\cite{aic21}. This challenge aims to boost the research on technologies applied to transportation systems and presents five different tracks. This project focuses on Track 3: City-Scale Multi-Camera Vehicle Tracking. The goal of this task is to track vehicles across multiple cameras intersections over a city. To train and test our models we use the track 3 dataset, referred to as CityFlowV2~\cite{cityflow}. This dataset contains sequences from 46 cameras spanning 16 intersections in a mid-sized city of U.S.A. More specifically we use 3 scenarios from the training dataset, formed by sequence 1, that captures a highway intersection, and sequences 3 and 4, that cover residential areas.

%% file: src/2_related_work.tex
\section{Related Work}
In this section, we provide a brief overview of existing works on different techniques that we used in Multi-Target Single-Camera (MTSC) and Multi-target Multi-Camera (MTMC) tracking. In both cases, the model must accurately detect objects in each frame and provide a consistent tracking of them, maintaining the same ID. With the dataset, a set of detections are given, obtained with well-known object detection algorithms such as YOLOv3~\cite{yolov3}, SSD512~\cite{ssd512} and Mask R-CNN~\cite{maskrcnn}. Our presented solution is a fine-tuned model of Faster R-CNN~\cite{faster} detector.

In terms of MTSC, we can find multiple state-of-the-art methods such as:
\begin{itemize}
   \item DeepSORT~\cite{sort}, an online method that combines deep learning features with Kalman-filter based tracking and the Hungarian algorithm.
   \item TC~\cite{TC}, which uses a fusion of visual and semantic features to cluster and associate data in each single camera view. Additionally, a histogram-based adaptive appearance model is introduced to learn long-term history of visual features for each vehicle target. 
   \item MOANA~\cite{moana}, which employs spatio-temporal data association using an adaptive appearance model to deal with identity switch caused by occlusion and similar appearance among nearby targets.
\end{itemize}

Finally, re-ID matches the objects across cameras on MTMC. Originally, most of the works focused on person re-ID, for example, Hermans \etal~\cite{hermans} with an implementation that used triplet loss with hard mining. These years, with an increasing interest on intelligent traffic management, vehicle re-ID has gained more attention and several proposals appeared. PROgressive Vehicle re-Identification (PROVID)~\cite{provid} presents a framework based on deep neural networks. It considers vehicle re-ID in two progressive procedures: coarse-to-fine search in the feature domain, and near-to-distant search in the physical space. Hongye Liu \etal~\cite{liu} propose a Deep Relative Distance Learning (DRDL) method which exploits a two-branch deep convolutional network to project raw vehicle images into an Euclidean space where distance can be directly used to measure the similarity of two vehicles.

%% file: src/3_methodology.tex
\begin{figure*}[t!]
    \centering
    \begin{tabular}{ccc}
    \includegraphics[width=0.27\textwidth]{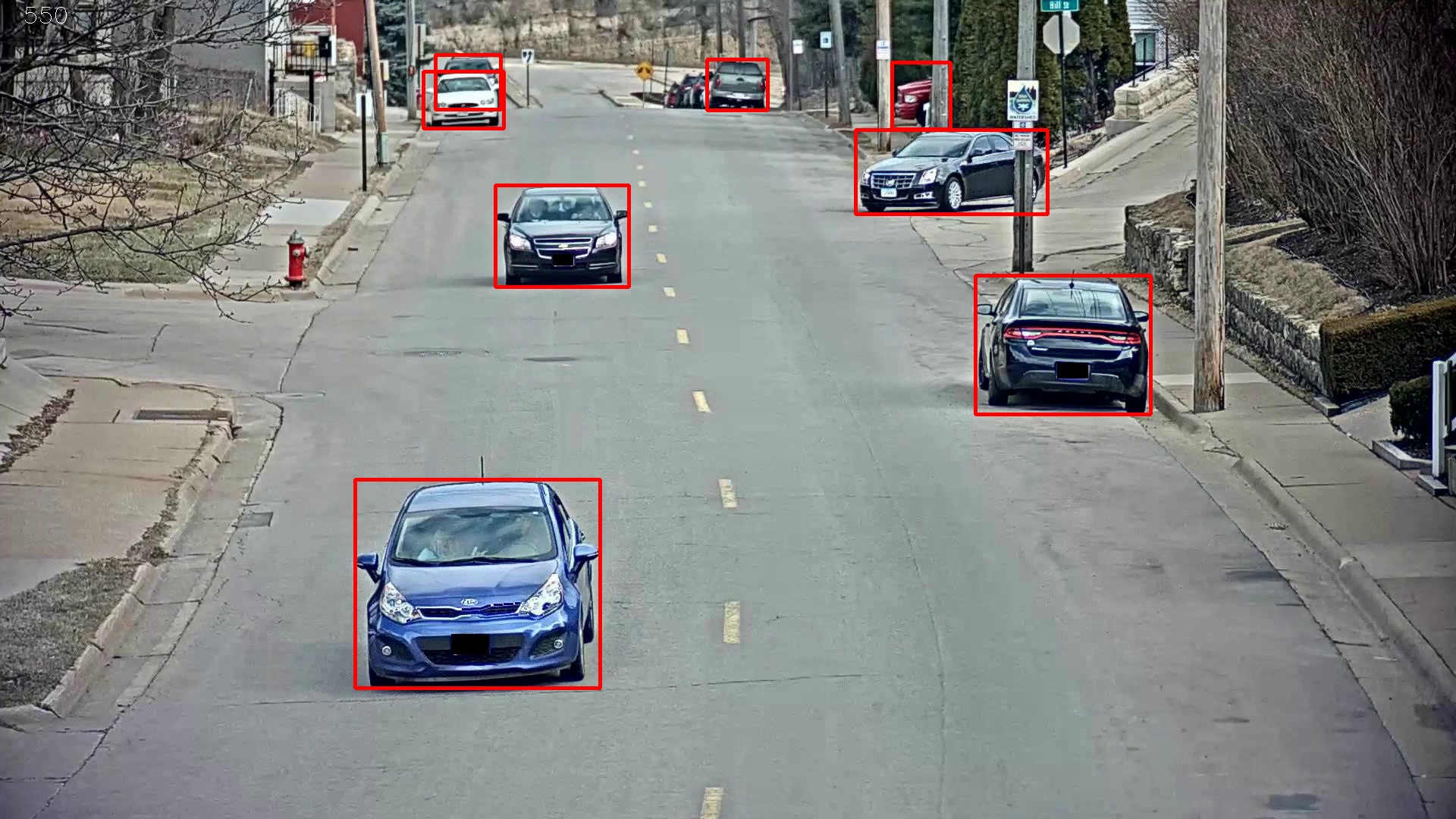} & 
    \includegraphics[width=0.27\textwidth]{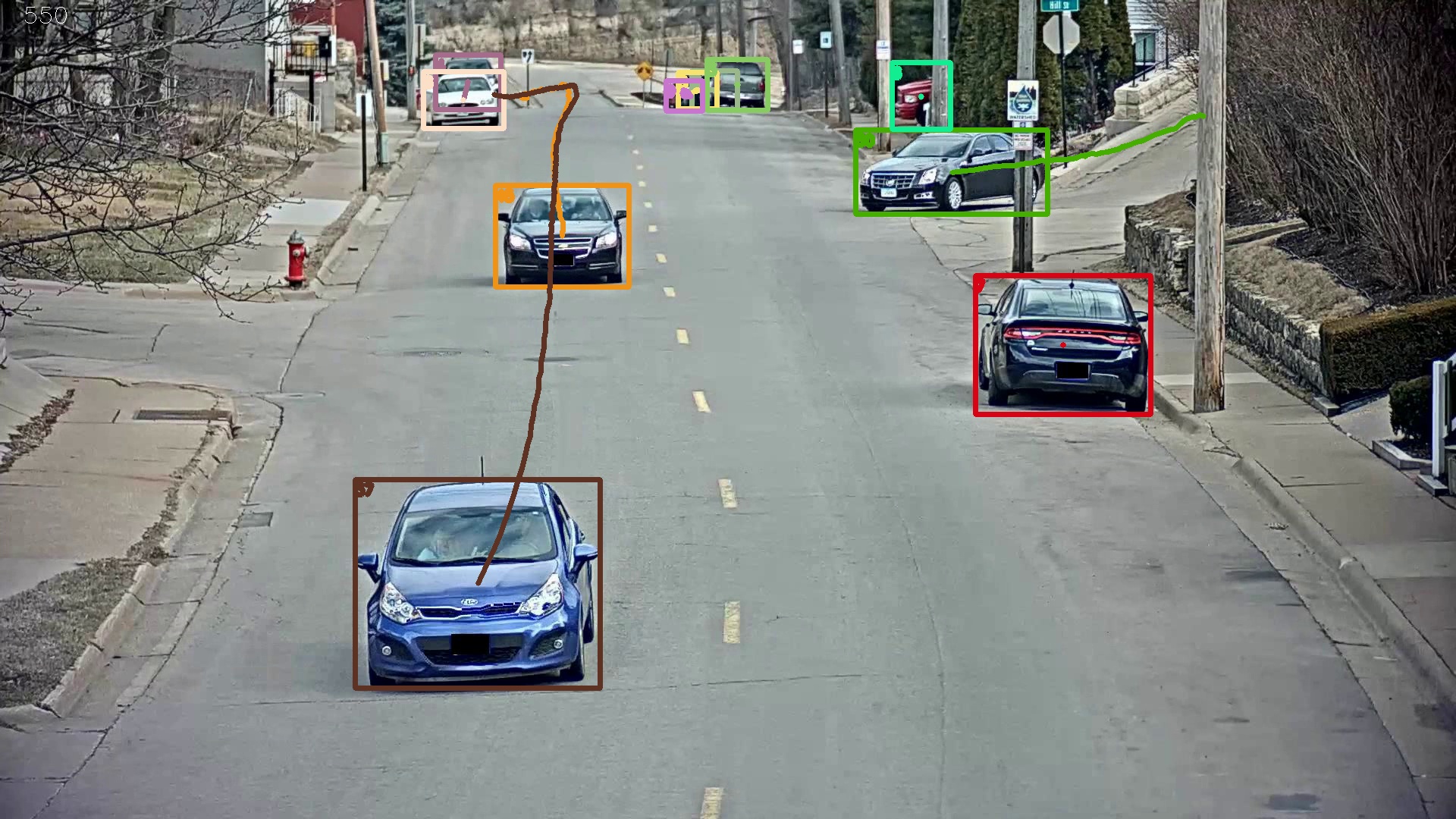} & 
    \includegraphics[width=0.27\textwidth]{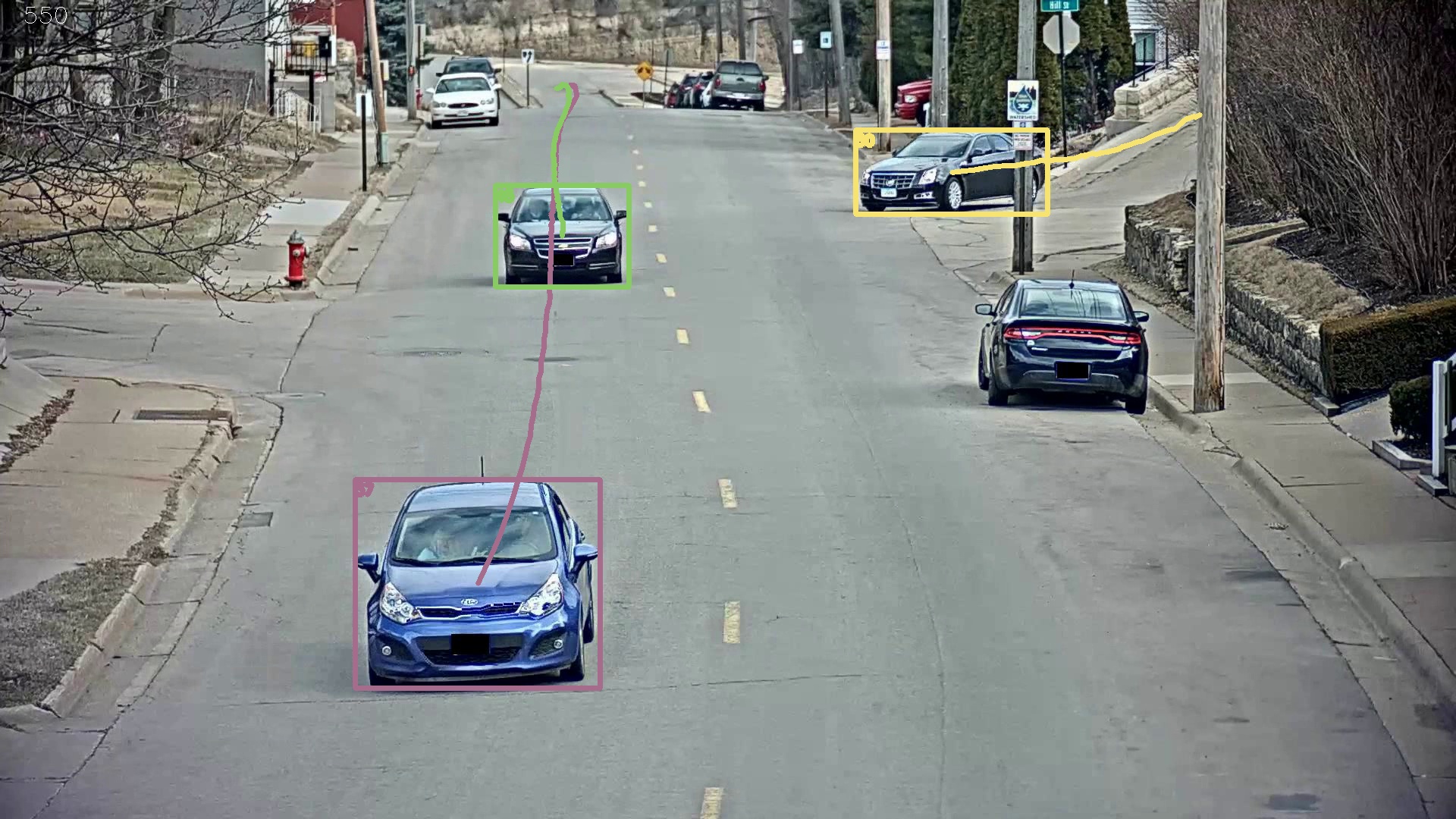} \\
    (a) & (b) & (c) \\
    \end{tabular}
    \caption{A single frame showing the results of the detections (a), tracking (b) and after post-processing (c).}
\label{fig:mtsc_method}
\end{figure*}

\section{Methodology}
This section is divided in two main parts: MTSC tracking (Sec. \ref{section:MTSC}) and MTMC tracking (Sec. \ref{section:MTMC}).

\subsection{Multi-Target Single-Camera Tracking}
\label{section:MTSC}

The main steps we carry out in order to implement MTSC tracking are:

 \begin{enumerate}
   \item Compute the \textbf{detections} of the cars for each frame separately.
   \item Use a \textbf{tracking method} to assign the same IDs to the same objects through the video.
   \item Apply a \textbf{post-processing} technique to remove small detections and parked cars to match the ground truth annotations.
 \end{enumerate}
 
These steps can be seen in Figure \ref{fig:mtsc_method}.

\subsubsection{Car detections}
\label{section:detections}
A first approach is to apply a background subtraction method, such as \textbf{MOG} \cite{MOG}, as these techniques only detect changing patterns on a video and thus only the moving objects will be detected. In this case, the main drawback is that not only cars will be detected, as there are other moving objects in the scene: bikes and pedestrians. To solve it, we apply a post-processing to keep detections with a similar aspect ratio and size of that of a car.

\input{tables/detections_results}

Deep Learning techniques are the state-of-the-art in object detection, so different models are compared in order to achieve the best results. Some detections are given with the dataset, but we wanted to fine-tune and optimize the hyper-parameters of one object detector. 

Therefore, we first obtain our own detections with the pre-trained \textbf{RetinaNet}~\cite{retina}, \textbf{YOLOv3}~\cite{yolov3}, \textbf{Mask R-CNN}~\cite{maskrcnn} and \textbf{Faster R-CNN}~\cite{faster} models. The results are presented in Table \ref{tab:table_object_detection} for the 75\% of frames obtained with camera 10 (sequence 3). Using a threshold confidence of 0.5, we can observe that the Faster R-CNN is the one that performs better, with a mAP$^{0.5}$ of 0.60.

Then, we fine-tune the pre-trained Faster-RCNN with the other 25\% of frames of camera 10. The following configuration is used: learning rate of 10$^{-3}$, SGD optimizer, momentum of 0.9 and weight decay at 10$^{-4}$. These parameters have been optimized using random 3-fold cross-validation technique, and the resulting mAP$^{0.5}$ is 0.97.

\subsubsection{Tracking method}
\label{section:tracking}
In order to maintain the ID of the same car along the whole video, we implemented two strategies: one based on maximum overlap and the other using a Kalman filter.

The \textbf{maximum overlap} tracking method is based on the Intersection over Union (IoU) metric. First, we assign new IDs to every Bounding Box (BB) detected in the first frame. Then, in the next frame, the IoU is computed between each new BB and the ones detected in the previous frame. The IDs of the previous detections are assigned to the BBs of the new frame with maximum IoU. When a new detection is not matched with any BB from the previous frame, it is considered a new object and then a new ID is generated.

The maximum overlap technique has a main drawback: the tracking of a car is lost when the car is not detected in a frame, and a new ID is assigned to it when it is detected again in the next frame. To tackle this problem, we use a \textbf{Kalman filter}, which allows us to predict the position of an object in the next frame, so the ID of a tracked car can be maintained along the sequence even if a detection is lost in a frame. We implemented the Kalman filter using the SORT~\cite{Bewley2016_sort} implementation.

Finally, we use \textbf{optical flow} to improve the tracking. The optical flow has been estimated with Lucas Kanade~\cite{lucas-kanade} algorithm, and it is used to shift the BBs of a frame to compensate the motion of the objects, and thus approximate them to the BBs of the next frame. 

\subsubsection{Post-processing}
\label{section:postprocessing}
Once the tracking is done, we adapt our results to compare them properly with the ground truth of AI City Challenge.

This dataset does not include parked cars, so the first step is to remove them. Specifically, for each detected car (with the same ID):

\begin{enumerate}
  \item We obtain the centers of its BBs for all the frames in which the car appears.
  \item We compute the variance of the centers.
  \item If the variance is below a threshold, in our case 50 pixels, the car is considered parked and thus removed.
\end{enumerate}

Also, small BBs (\eg cars that are far away) are not annotated, so we keep only the BBs with a minimum width of 80 pixels and a minimum height of 60 pixels.

\begin{figure}[t!]
    \centering
    \begin{tabular}{ccc}
    \includegraphics[width=0.20\linewidth]{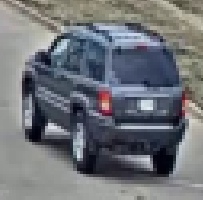} & 
    \includegraphics[width=0.20\linewidth]{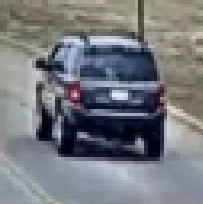} & 
    \includegraphics[width=0.20\linewidth]{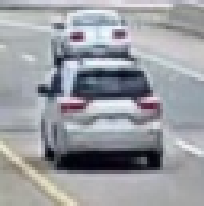} \\
    (a) & (b) & (c) \\
    \end{tabular}
    \caption{An example of a triplet from the mining function: anchor (a), positive pair (b) and negative pair (c).}
\label{fig:triplet}
\end{figure}

\subsection{Multi-Target Multi-Camera Tracking}
\label{section:MTMC}
In order to solve the problem of keeping the same ID across all cameras, we use Pytorch Metric Learning, an open library~\cite{musgrave2020pytorch} that allows us to include metric learning in our pipeline with the following steps:

\begin{enumerate}
  \item Create a dataset containing images of all cars from different cameras with their label ID.
  \item Train a siamese network to cluster the instances.
  \item Create a re-identification algorithm using the trained network to find matches of a given car instance.
\end{enumerate}

\subsubsection{Clustering with a Siamese Network}

First of all, we obtain the car patches using the ground truth annotations from all cameras of sequences 1 and 4. These patches are used to train a siamese network: a ResNet-18~\cite{resnet18} with the last layer removed and replaced by an embedding layer of dimension 256. The network is trained with triplet loss, and mining is applied to create hard positives and negatives pairs from the anchor image, as presented in Fig.\ref{fig:triplet}. 

The output of the siamese network is a clustering of the embedding vectors, which is computed with a cosine similarity as distance function. Finally, we design a system to re-assign the IDs of cars across different cameras using this embedding space.

\subsubsection{Re-ID system}
\label{section:REID}
To explain our algorithm we use two cameras: cam1 and cam2. For each camera, we've tracked each car separately, and we crop them to obtain car patches with their corresponding IDs. One camera is defined as the reference camera, so all the IDs of its cars are maintained: in this case cam1. Then, we take P patches from a car of cam2, and compare them (independently) with N patches of each car of cam1, in an all vs. all approach. This comparison is performed using the siamese network, which provides a metric distance between patches. If this distance is below a threshold, the two patches are considered a match. We count the number of matches that each car of cam1 has, and the ID of the car with more matches is assigned to the car of cam2.

Using more than one patch for each car ensures that the comparison is more robust, as it might be that some patches of a car are not representative enough (\eg due to occlusions). These P (or N) patches are not selected randomly: if a car appears in M (consecutive) frames in a camera, we take a patch for each $M/P$ (or $M/N$) frames. This way, the patches are more variate and thus useful for the comparison.

To extrapolate this system to the six cameras of sequence 3, we propose a cascade scheme. First, two cameras are compared taking one as a reference. Once the re-id is done, these two cameras are combined and taken as a reference, to be compared with another one. This is done until all cameras have been re-identified. This scheme has been implemented in a random order of cameras, but we could have performed a spatio-temporal analysis to know which cameras to look for at each moment. However, the cameras are not well synchronized, as some frames are lost in the middle of the sequence. Moreover, by doing this analysis we would have over fitted the system to this specific sequence, as this would need a different configuration if the cameras were different.

%% file: tables/detections_results.tex
\begin{table}[t!]
\begin{center}
\begin{tabular}{|c|c|c|}
\hline
\textbf{Model} & \textbf{mAP$^{0.5}$} & \textbf{Inference time (s/img)} \\ \hline\hline
RetinaNet      & 0.53       & 0.098                         \\ \hline
YOLOv3         & 0.54       & 0.031                         \\ \hline
Mask R-CNN     & 0.48       & 0.516                         \\ \hline
Faster R-CNN   & 0.60       & 0.105                          \\ \hline
\end{tabular}
\end{center}
\caption{Comparison of mAP$^{0.5}$ and inference time between object detection models applied to our case.}
\label{tab:table_object_detection}
\end{table}

%% file: src/4_evaluation.tex
\begin{figure}[t!]
    \centering
    \begin{tabular}{cc}
         \includegraphics[width=0.45\linewidth]{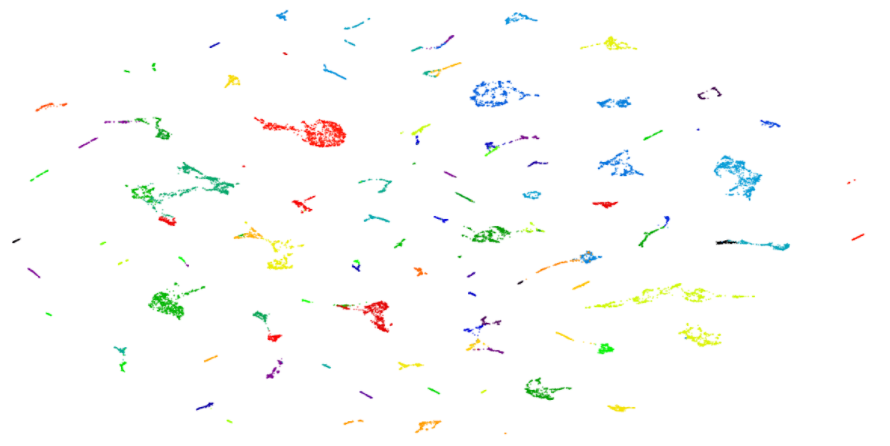} & \includegraphics[width=0.45\linewidth]{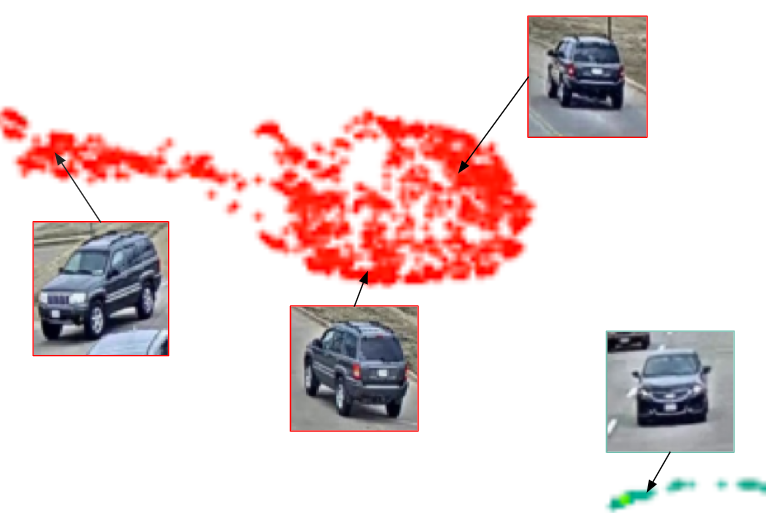} \\
        (a) & (b) \\
    \end{tabular}
    \caption{Clustering in the embedding space (a), and an interpretation of a zoom of a given cluster (b).}
\label{fig:mtmc_embedding}
\end{figure}

\input{tables/mtsc_results}

\begin{figure}[t!]
\begin{center}
\includegraphics[width=\linewidth]{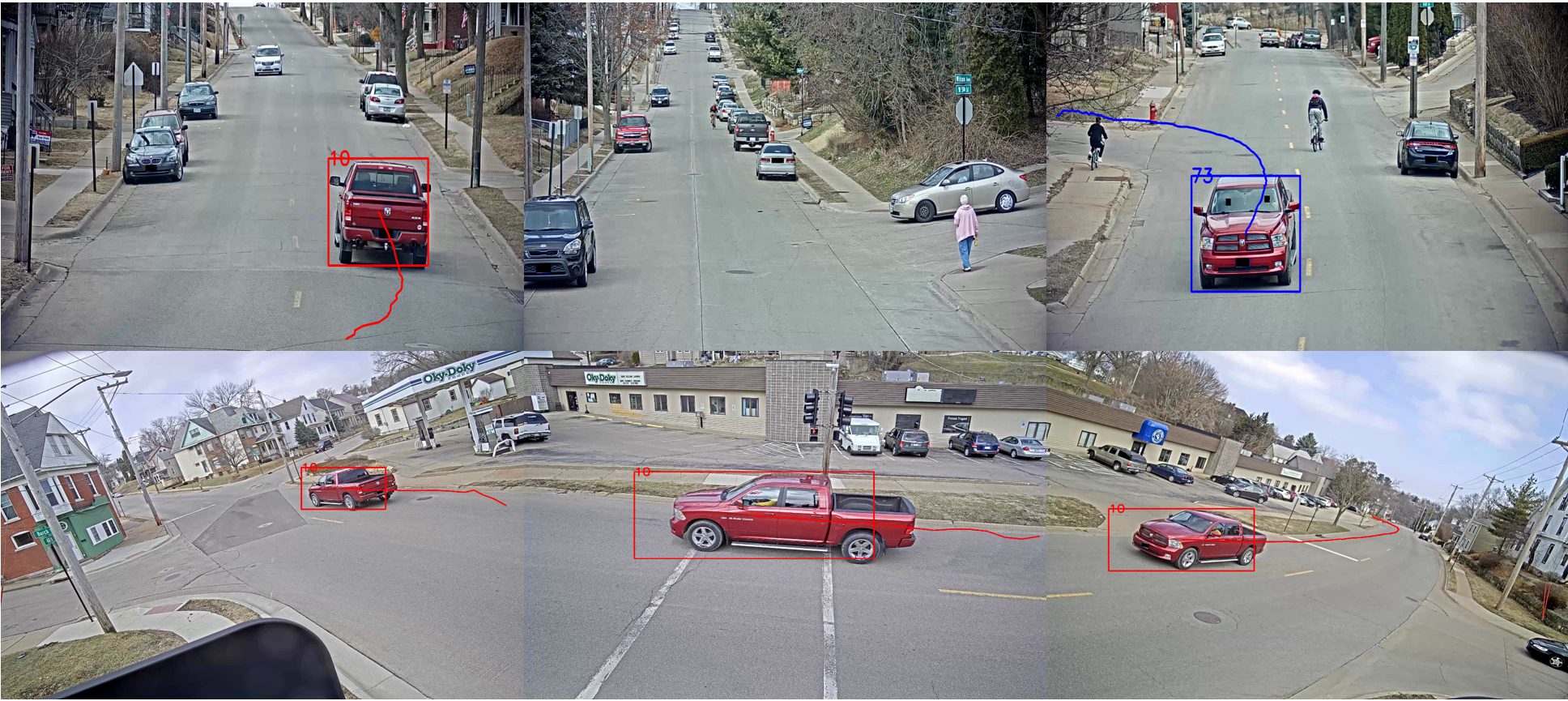}
\end{center}
   \caption{Vehicle correctly tracked across four cameras (red) and erroneously in one (blue).}
\label{fig:mtmc_qualitative}
\end{figure}

\section{Evaluation}

\subsection{MTSC Tracking}
\label{section:MTSC_results}

To evaluate the MTSC tracking methods we compute the IDF1 separately for all the cameras (10-15) of sequence 3 and the resulting average. A comparison between the implemented algorithms is presented in Table \ref{tab:mtsc_comparison}. The best result is obtained using the fine-tuned Faster R-CNN object detector, with maximum overlap as the tracking method (average IDF1 of 0.72). We also observe that the performance of both tracking methods is similar.

There are some inconsistencies in the results of cameras 12 and 15, which we identify as problems with the ground truth. Specifically, some cars that are moving (and thus tracked) do not appear in more than one camera, so they are not included in the ground truth. On the other hand, there are cars that are parked (and thus not tracked), but they are annotated the whole sequence because at some point they start moving, which decreases the precision of our system. These situations lead to some unfair comparisons, as for example Faster R-CNN performs well in camera 12 because the detections with this network are better suited for this ground truth, while the results of the other methods are poor.

\input{tables/mtmc_results}

\subsection{MTMC Tracking}
\label{section:MTMC_results}
The siamese network is able to cluster in a 256-d embedding space the car instances of the dataset created with sequences 1 and 4, as observed in Fig.~\ref{fig:mtmc_embedding}. After fine-tuning the network, the highest test accuracy obtained when matching car instances in sequence 3 is 0.43.

In Table \ref{tab:mtmc_results} we present the MTMC results in sequence 3, using our final configuration: threshold to consider a match of $0.6$, $N = 3$ and $P = 4$. As expected, MTMC performs worse in terms of IDF1 score than MTSC. This makes sense if we consider the augmented difficulty of the task, and the low accuracy obtained with the siamese network. Nonetheless, in the qualitative results presented in Fig.~\ref{fig:mtmc_qualitative}, we are able to track correctly a vehicle with the same id across 4 out of 5 cameras.

%% file: tables/mtsc_results.tex
\begin{table*}[t!]
\begin{center}
\begin{tabular}{|c|c|cccccc|c|}
\hline
\textbf{Detection Method} & \textbf{Tracking Method} & \textbf{c10}  & \textbf{c11} & \textbf{c12}  & \textbf{c13}  & \textbf{c14}  & \textbf{c15}  & \textbf{Average} \\ 
\hline\hline
\multirow{2}{*}{MOG} & M. Overlap & 0.79  & 0.85  & 0.12  & 0.79   & 0.63  & 0.09 & 0.55 \\ \cline{2-9} 
& Kalman & 0.84 & 0.74 & 0.11 & 0.67 & 0.73 & 0.08 & 0.53 \\ 
\hline
\multirow{2}{*}{Faster R-CNN (FT)} & M. Overlap   & 0.88 & 0.73 & 0.92 & \textbf{0.86} & \textbf{0.87} & 0.06  & \textbf{0.72} \\ \cline{2-9} 
& Kalman & \textbf{0.90} & 0.69 & \textbf{0.95} & 0.79  & 0.85  & 0.05  & 0.71  \\ 
\hline
\multirow{2}{*}{Mask R-CNN} & M. Overlap  & 0.88 & 0.69 & 0.11  & 0.85 & 0.73 & 0.14 & 0.57 \\ 
\cline{2-9} 
& Kalman  & 0.86 & 0.47 & 0.11 & 0.73 & 0.80 & 0.14 & 0.52 \\ 
\hline 
\multirow{2}{*}{YOLOv3} & M. Overlap & 0.87 & 0.69 & 0.07 & 0.72 & 0.61 & 0.14  & 0.52 \\ 
\cline{2-9} 
& Kalman & 0.89 & 0.61 & 0.06 & 0.64 & 0.77 & 0.14 & 0.52 \\ 
\hline 
\multirow{2}{*}{SSD512} & M. Overlap  & 0.88 & 0.62  & 0.10 & 0.65 & 0.57   & \textbf{0.23} & 0.51 \\
\cline{2-9}
& Kalman   & 0.87 & 0.60  & 0.09  & 0.58  & 0.67  & 0.19          & 0.50 \\ 
\hline
\end{tabular}
\end{center}
\caption{Multi Target Single Camera methods comparison for all cameras.}
\label{tab:mtsc_comparison}
\end{table*}

%% file: tables/mtmc_results.tex
\begin{table}[t!]
\begin{center}
\begin{tabular}{|c|ccc|cc|}
\hline
Sequence & IDF1 & IDP & IDR & Precision & Recall \\
\hline\hline
3 & 0.59 & 0.56 & 0.65 & 0.79 & 0.93 \\
\hline
\end{tabular}
\end{center}
\caption{Results obtained with MTMC tracking and object detection on sequence 3.}
\label{tab:mtmc_results}
\end{table}

%% file: src/5_conclusions.tex
\section{Conclusions}
We have implemented a pipeline able to track multiple vehicles successfully on a single camera. The best results are obtained with Faster R-CNN, which has been fine-tuned and optimized for this specific dataset. Both tracking methods, maximum overlap and Kalman, worked similarly, being the former slightly better. However, the quantitative results obtained were low due to inconsistencies with the annotations. Then, we extended the pipeline to re-identify vehicles and track them across multiple cameras, obtaining worse results than in the single camera case. In the MTMC scenario, we rely on the performance of the siamese network, and its low accuracy impairs the overall re-ID system.

Finally, the lack of temporal information converts our approach in an ill-posed problem, as it is impossible to obtain a solution in which two vehicles with the same colour and model are differentiated without a time reference. The lack of synchronization between cameras makes it hard to design a model that solves this problem without over fitting to the used sequence. It would require a manual synchronization for each specific sequence of frames and objects to be tracked.